\documentclass[conference]{IEEEtran}
\IEEEoverridecommandlockouts

\usepackage{cite}
\usepackage{amsmath,amssymb}
\usepackage{graphicx}
\usepackage{booktabs}
\usepackage{url}
\usepackage[hidelinks]{hyperref}

\begin{document}

\title{How Visible Are Silent Manipulation Failures?\\
An Observability Study of False-Success Detection\\
in Simulated Robot Episodes}

\author{\IEEEauthorblockN{Aarav Bedi}
\IEEEauthorblockA{Department of Mechanical Engineering\\
University of California, Berkeley\\
Berkeley, California, USA\\
aaravbedi@berkeley.edu}}

\maketitle

\begin{abstract}
Imitation-learning policies for robot manipulation inherit the quality of the
success labels attached to their training episodes, and those labels are usually
produced by the robot's own success check. A particularly damaging error is the
false success: an episode the robot logs as a success when the task outcome was
actually wrong. We ask a narrow but practical question about these episodes. Once
an episode has already been flagged as a success, how much of the information
needed to overturn that label is present in proprioception, and how much requires
vision? We build a simulated testbed on two bimanual ALOHA tasks, induce failures
through environment perturbations rather than label edits, label every episode by
privileged simulator state that the detector never sees, and keep only episodes the
robot flagged as successful. We then compare detectors restricted to proprioception
against a vision-based detector. We find that recoverability spans a wide range: in
cube transfer the false successes are almost fully recoverable from joint data
alone, while in peg insertion proprioception recovers only part of them and a
vision detector closes most of the gap. We also show that the proprioceptive
separability we measure rests on velocity differences far below any realistic sensor
noise floor, so it is best read as an optimistic upper bound that a noiseless
simulator inflates. We release the generation and evaluation pipeline.
\end{abstract}


\section{Introduction}
Manipulation policies are increasingly trained by imitation from large demonstration
corpora~\cite{oxe,bridge,droid,lerobot}, and every one of those corpora rests on the
assumption that each episode carries a correct success or failure label. In practice
the label is generated automatically, by the robot's own success detector or a
scripted end-of-trajectory check. When the check fires on an episode that did not
actually accomplish the task, the resulting false success enters the training set as
a positive example, and the policy has no way to know that the behavior it is
imitating ended badly.

False successes are harder to find than ordinary execution failures. A stalled joint
or a violent jerk leaves an obvious trace in the sensed trajectory; a cube placed two
centimeters off target, or a peg that looks seated but is not, may leave almost none.
The practical question for anyone trying to clean a training set is therefore not
whether a failure occurred, but which sensing channel carries enough signal to
recover it after the robot has already declared victory.

This paper studies that question directly in simulation. We deliberately avoid the
shortcut of injecting failures as label edits, because a detector trained on such
data learns to recover the injection rule rather than a real failure signature. We
instead run a scripted controller in a physics simulator, perturb the environment so
that genuine misses occur, and read the true outcome from privileged simulator state
that the detector is never given. We keep only the episodes the robot flagged as
successful, so that the task is exactly the one a curation pipeline faces: separate
the real successes from the false ones using only observable signals.

Our contribution is an observability analysis rather than a new detection method. We
report, for two manipulation tasks, how far a proprioception-only detector and a
vision detector get on the false-success problem, and we are explicit about where the
proprioceptive signal comes from and how fragile it is. Two findings stand out. First,
recoverability is not uniform: transfer failures are almost fully visible to joint
data, while insertion failures are only partly visible and need vision to be caught
reliably. Second, the proprioceptive separability we measure is driven by velocity
differences on the order of one part in a thousand, which a noiseless simulator
resolves perfectly but a real robot's sensors would not. We therefore treat the
proprioception numbers as an upper bound. The pipeline is released so the analysis can
be extended to other tasks, policies, and modalities.

\section{Related Work}
\textbf{Manipulation datasets.} Large demonstration corpora have driven recent
progress in robot learning. Open X-Embodiment~\cite{oxe} aggregates data across many
robots and labs, BridgeData V2~\cite{bridge} and DROID~\cite{droid} provide large
single-embodiment collections, and LeRobot~\cite{lerobot} standardizes the data format
and supplies widely used reference datasets. These efforts target scale and coverage
and do not provide a labeled measure of annotation quality.

\textbf{Failure and anomaly detection.} Detecting execution failures from
proprioception and force is well studied, with residual-monitoring and learned
sequence-model approaches~\cite{park,fino}. These methods work precisely because the
failures they target leave a signature in the sensed trajectory. The case where the
trajectory looks nominal but the outcome is wrong has received much less attention.

\textbf{Success detection and reward models.} Vision-based success classifiers and
learned reward models judge task outcome from images~\cite{vlmsuccess}, and are the
natural tool for catching false successes. What has been missing is a controlled
measurement of how much of the false-success signal lives in proprioception versus
vision once the robot has already reported success. The simulated ALOHA tasks and
scripted policies we build on come from the ACT line of work~\cite{act}.

\section{Testbed Design}
Both tasks derive from the bimanual ALOHA simulator~\cite{act,lerobot}: a cube transfer
task and a peg insertion task. We use the scripted end-effector controllers that ship
with the simulator so that nominal behavior is competent and reproducible, and we draw
the true outcome from the simulator's privileged object state.

\textbf{Inducing failures.} We do not edit labels. For each episode we apply randomized
environment perturbations (Table~\ref{tab:data}): a small planar jitter and yaw
perturbation of the manipulated object, and randomized object friction. The scripted
controller then runs to completion. Some perturbed episodes still succeed and some do
not; the failures are a consequence of the physics, not of an annotation.

\textbf{Two labels, kept apart.} Each episode carries two distinct labels. The
\emph{robot flag} is a cheap success heuristic computed from proprioception only,
standing in for the telemetry-based check a real robot would use. The
\emph{ground-truth outcome} is read from privileged simulator state, the object pose
relative to its target, which the detector never observes. A false success is an
episode where the robot flag fired but the ground-truth outcome is a failure. We retain
only flagged-success episodes, so the dataset is exactly the pool a curation step would
inspect.

\textbf{Failure modes.} Modes are assigned after the fact from the ground-truth state,
not injected. In transfer the failures are dominated by a misplaced cube, with a small
number of drops; in insertion the single observed mode is a peg that ends up not seated
near the socket. We report whichever modes occur rather than forcing a fixed taxonomy.

\begin{table}[t]
\centering
\caption{Testbed composition (flagged-success episodes only, seed 42)}
\label{tab:data}
\begin{tabular}{lcc}
\toprule
& Transfer & Insertion \\
\midrule
Flagged-success episodes & 500 & 500 \\
True successes & 262 & 341 \\
False successes & 238 & 159 \\
False-success rate & 47.6\% & 31.8\% \\
\midrule
Object planar jitter & 1.5\,cm & 0.5\,cm \\
Object yaw jitter & $15^\circ$ & $5^\circ$ \\
Object friction range & 0.15--0.9 & 0.15--0.9 \\
Added observation noise & none & none \\
\bottomrule
\end{tabular}
\end{table}

\section{Evaluation Protocol}
The task is stated as follows: given an episode already flagged as a success, predict
whether it actually succeeded. The headline metric is \emph{false-success recall}, the
fraction of truly failed episodes that the detector flags, because that is the quantity
a curation step cares about and because overall accuracy is misleading when one class
dominates.

We compare two detectors. \textbf{Baseline A (proprioception)} uses summary statistics
of the twelve arm joints' velocity over the full episode: three global statistics (the
root-mean-square, peak, and standard deviation of the per-step joint-velocity
magnitude) together with the root-mean-square, peak, and standard deviation of each
individual joint's velocity, for thirty-nine features in total. No privileged state,
perturbation parameter, or failure-mode label is used as a feature. \textbf{Baseline C
(vision)} uses three features read from the rendered camera image of the final state:
the normalized image centroid $(c_x, c_y)$ and the pixel area of the manipulated
object, obtained by color-thresholding the rendered frame. These come from the image
alone; the detector never reads object pose from simulator state. Both detectors are
gradient-boosted tree
classifiers with class balancing, trained on a fixed split and evaluated on a held-out
test set of 150 episodes. The seed is fixed at 42.

As a diagnostic that the proprioceptive features are not trivially separating the
classes, we report Cohen's $d$ between true and false successes for every feature and
for each trajectory window. Large $d$ indicates that the failure is, in fact, visible
to proprioception.

\section{Results}
\begin{figure}[t]
\centering
\includegraphics[width=0.9\linewidth]{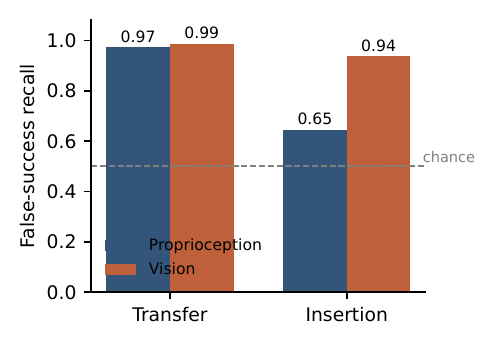}
\caption{False-success recall by sensing modality. Transfer failures are almost fully
recoverable from proprioception; insertion failures need vision to close the gap.}
\label{fig:recall}
\end{figure}

\textbf{Recoverability differs sharply by task.} Fig.~\ref{fig:recall} and
Table~\ref{tab:baselines} give the headline numbers. In transfer, proprioception alone
recovers 97\% of the false successes and vision adds little, because the failures
change the load the arm carries and so are written plainly into the joint trajectory.
In insertion, proprioception recovers only 65\%, while the vision detector reaches 94\%.
Insertion is where the modality gap is real.

\begin{table}[t]
\centering
\caption{Detector metrics on the held-out test set (150 episodes)}
\label{tab:baselines}
\begin{tabular}{llcccc}
\toprule
Task & Detector & Acc. & FS recall & TS recall & Macro F1 \\
\midrule
Transfer & Proprioception & 0.973 & 0.972 & 0.975 & 0.973 \\
Transfer & Vision & 0.987 & 0.986 & 0.987 & 0.987 \\
Insertion & Proprioception & 0.760 & 0.646 & 0.814 & 0.727 \\
Insertion & Vision & 0.973 & 0.938 & 0.990 & 0.969 \\
\bottomrule
\end{tabular}
\end{table}

\begin{figure}[t]
\centering
\includegraphics[width=0.9\linewidth]{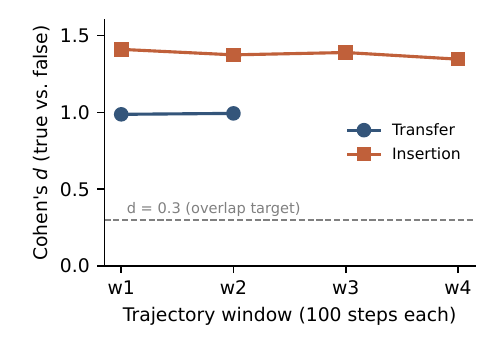}
\caption{Per-window Cohen's $d$ between true and false successes. The separating signal
is present across the whole trajectory in both tasks, well above the overlap target.}
\label{fig:window}
\end{figure}

\begin{figure*}[t]
\centering
\includegraphics[width=0.95\linewidth]{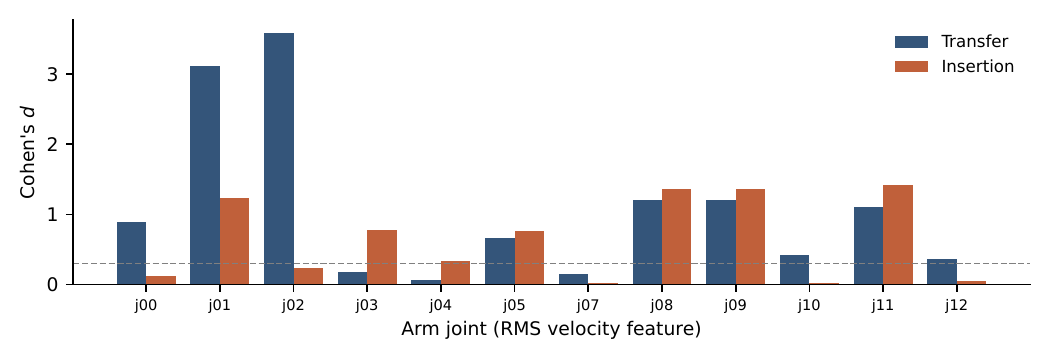}
\caption{Per-joint Cohen's $d$ of the RMS velocity feature. The proprioceptive signal
that distinguishes false successes is concentrated in a few joints, and is much
stronger in transfer than in insertion.}
\label{fig:joint}
\end{figure*}

\textbf{The proprioceptive signal is present in every window.}
Fig.~\ref{fig:window} shows Cohen's $d$ between true and false successes within each
100-step window. For transfer it sits near $1.0$ throughout, and for insertion above
$1.3$ in every window. There is no segment of the trajectory in which the two classes
become indistinguishable, which rules out the possibility that a particular feature
window is doing the separating. Fig.~\ref{fig:joint} shows the signal is concentrated
in a few joints rather than spread evenly.

\textbf{But the signal is tiny in absolute terms.} Table~\ref{tab:tiny} pairs each
window's Cohen's $d$ with the actual difference between the class means. In transfer's
first window the mean feature differs by less than $0.001$ in normalized units, yet
$d\approx 0.99$. The large effect size comes almost entirely from near-zero
within-class variance, which the deterministic simulator produces and a real sensor
would not. The proprioceptive separability is therefore real in this simulator and
fragile outside it.

\begin{table}[t]
\centering
\caption{Per-window separability vs. absolute mean difference (transfer)}
\label{tab:tiny}
\begin{tabular}{lccc}
\toprule
Window & mean (true) & mean (false) & Cohen's $d$ \\
\midrule
t000--100 & 0.8598 & 0.8608 & 0.987 \\
t100--200 & 0.2002 & 0.1929 & 0.993 \\
\bottomrule
\end{tabular}
\end{table}

\section{Discussion and Limitations}
The headline of this study is a caution as much as a result. A clean reading of the
numbers would say that false successes are mostly recoverable from proprioception, but
that reading would be wrong for two reasons, and stating them plainly is the point of
the paper.

\textbf{Transfer is essentially physical-failure detection.} The transfer failures we
induced change the mass the arm carries, so they show up in the joint trajectory and a
proprioception detector catches them at ceiling. These are not the silent failures the
problem is about; they are ordinary execution failures that happen to be mislabeled as
successes by a loose flag. The only task here that exhibits a genuine proprioception
versus vision gap is insertion.

\textbf{The proprioceptive separability is a simulation artifact.} As
Table~\ref{tab:tiny} shows, the velocity differences that separate the classes are on
the order of one part in a thousand. We added no observation noise. A real robot's
encoder and velocity estimates carry noise that would bury a signal this small, so the
0.97 and 0.65 numbers should be read as optimistic upper bounds rather than as
expected field performance. This makes the case for vision stronger, not weaker.

\textbf{The insertion signal is partly a policy artifact.} The scripted controller
steers toward the true peg pose, so when the peg is jittered the arm's trajectory
adapts, and that adaptation correlates with eventual success. A learned, closed-loop,
or fixed open-loop policy would distribute this signal differently. The insertion
proprioception number is therefore tied to the oracle controller and should not be
read as a property of the task alone.

\textbf{Scope.} The study covers two simulated tasks, a single random seed, 500
flagged-success episodes per task, and an oracle scripted policy. The failures that
arise are dominated by one or two modes per task rather than a broad taxonomy. We
report a force-augmented detector in the released code but omit it here because it did
not differ measurably from the proprioception detector, and we did not want to present
an uninformative comparison as a finding.

\textbf{Future work.} The natural next steps are multiple seeds with confidence
intervals, learned policies in place of the oracle controller, calibrated
proprioceptive sensor noise so that detectability reflects realistic conditions, and a
failure-induction mechanism that perturbs the object relative to the gripper without
disturbing the arm, which is the regime in which a truly proprioception-invisible
failure would arise. Real-robot episodes are the eventual goal.

\section{Conclusion}
We studied how recoverable false-success manipulation failures are from different
sensing modalities, using a simulated testbed in which failures arise from physics
rather than label edits and ground truth is hidden from the detector. Recoverability
ranges from near-complete for transfer, where the failures perturb the arm's own
dynamics, to partial for insertion, where vision is needed to close the gap. We further
show that the proprioceptive separability we measure rests on sub-sensor-noise velocity
differences and on the oracle controller, so it overstates what a real system would
recover. The honest takeaway for training-data curation is that catching silent
manipulation failures generally requires exteroception, and that clean-simulation
proprioception results on this problem should be treated with care.

\section*{Data and Code Availability}
The episode-generation and evaluation pipeline, including the feature extractors and the
summary used to produce all figures and tables in this paper, is available at
\url{https://github.com/aaravbedi/Silent-Manipulation-Failures}.

\section*{Acknowledgment}
This work builds on open-source benchmarking tools and datasets released by HaptalAI
(\url{https://huggingface.co/HaptalAI}). The author conducted informal consultations
with PhD students and simulation engineers across industry and academia, which helped
shape the problem framing and failure taxonomy. AI assistance (Anthropic's Claude) was used for editing and formatting the written manuscript. All simulation
code, experiments, and analysis were implemented and executed by the author, who takes
full responsibility for all results and conclusions.

\end{document}